\documentclass[journal]{IEEEtran}
\usepackage{graphicx}
\usepackage{amsmath}
\usepackage{amssymb}
\usepackage{booktabs}
\usepackage{mathtools}
\usepackage{cite}
\usepackage{adjustbox}
\usepackage{multicol}
\usepackage{multirow}
\usepackage{pifont}
\usepackage{dsfont}
\usepackage{array}
\usepackage{dblfloatfix}
\usepackage{enumitem}
\usepackage{times}
\usepackage{epsfig}
\usepackage{balance}
\usepackage{url}
\usepackage[pagebackref=true,breaklinks=true,letterpaper=true,colorlinks,bookmarks=false]{hyperref}

\ifCLASSINFOpdf
\else
\fi

\begin{document}
\title{Prototype as Query for Few Shot Semantic Segmentation}

\author{Leilei Cao,~\IEEEmembership{Member,~IEEE,}
        Yibo Guo,
        Ye Yuan
        and Qiangguo Jin,~\IEEEmembership{Member,~IEEE}
\thanks{Manuscript received xxxx, 2022; revised xxxx, 2023.}
\thanks{L. Cao and Q. Jin are with the School
of Software, Northwestern Polytechnical University, Shannxi, China.(E-mail: caoleilei@nwpu.edu.cn)}
\thanks{Y. Guo is with Suzhou Zhito Technology Co., Ltd, Jiangsu, China.}
\thanks{Y. Yuan is with Ping An Property \& Casualty Insurance Company of China Ltd, Shenzhen, China.}
}

\markboth{IEEE Transactions on Cybernetics}%
{Shell \MakeLowercase{\textit{et al.}}: Bare Demo of IEEEtran.cls for IEEE Journals}

\maketitle

\begin{abstract}
Few-shot Semantic Segmentation (FSS) was proposed to segment unseen classes in a query image, referring to only a few annotated examples named support images. One of the characteristics of FSS is spatial inconsistency between query and support targets, e.g., texture or appearance. This greatly challenges the generalization ability of methods for FSS, which requires to effectively exploit the dependency of the query image and the support examples. Most existing methods abstracted support features into prototype vectors and implemented the interaction with query features using cosine similarity or feature concatenation. However, this simple interaction may not capture spatial details in query features. To alleviate this limitation, a few methods utilized all pixel-wise support information via computing the pixel-wise correlations between paired query and support features implemented with the attention mechanism of Transformer. These approaches suffer from heavy computation on the dot-product attention between all pixels of support and query features. In this paper, we propose a simple yet effective framework built upon Transformer termed as ProtoFormer to fully capture spatial details in query features. It views the abstracted prototype of the target class in support features as Query and the query features as Key and Value embeddings, which are input to the Transformer decoder. In this way, the spatial details can be better captured and the semantic features of target class in the query image can be focused. The output of the Transformer-based module can be viewed as semantic-aware dynamic kernels to filter out the segmentation mask from the enriched query features. Extensive experiments on PASCAL-$5^{i}$ and COCO-$20^{i}$ show that our ProtoFormer significantly advances the state-of-the-art methods.
\end{abstract}


\IEEEpeerreviewmaketitle

\section{Introduction}
\IEEEPARstart{W}{ith} the development of deep Convolutional Neural Networks (CNNs), semantic segmentation, a fundamental task in computer vision, has been made remarkable progress\cite{deeplab,deeplabv3,long2015,fcn,cheng2021maskformer,xie2021segformer}. However, the good performance of methods on semantic segmentation relies heavily on large-scale datasets with annotations\cite{zhou2017scene,cocostuff}, which consumes amount of time and labor for collecting images and densely annotating. To alleviate the data-hungry in real-world scenarios, Few-shot Semantic Segmentation (FSS)\cite{shaban,prototypenet} was proposed to segment unseen object classes in a query image, referring to only a few annotated examples named support images which contain the target class. 

\begin{figure}[t]
  \centering
  \includegraphics[width=1.0\linewidth]{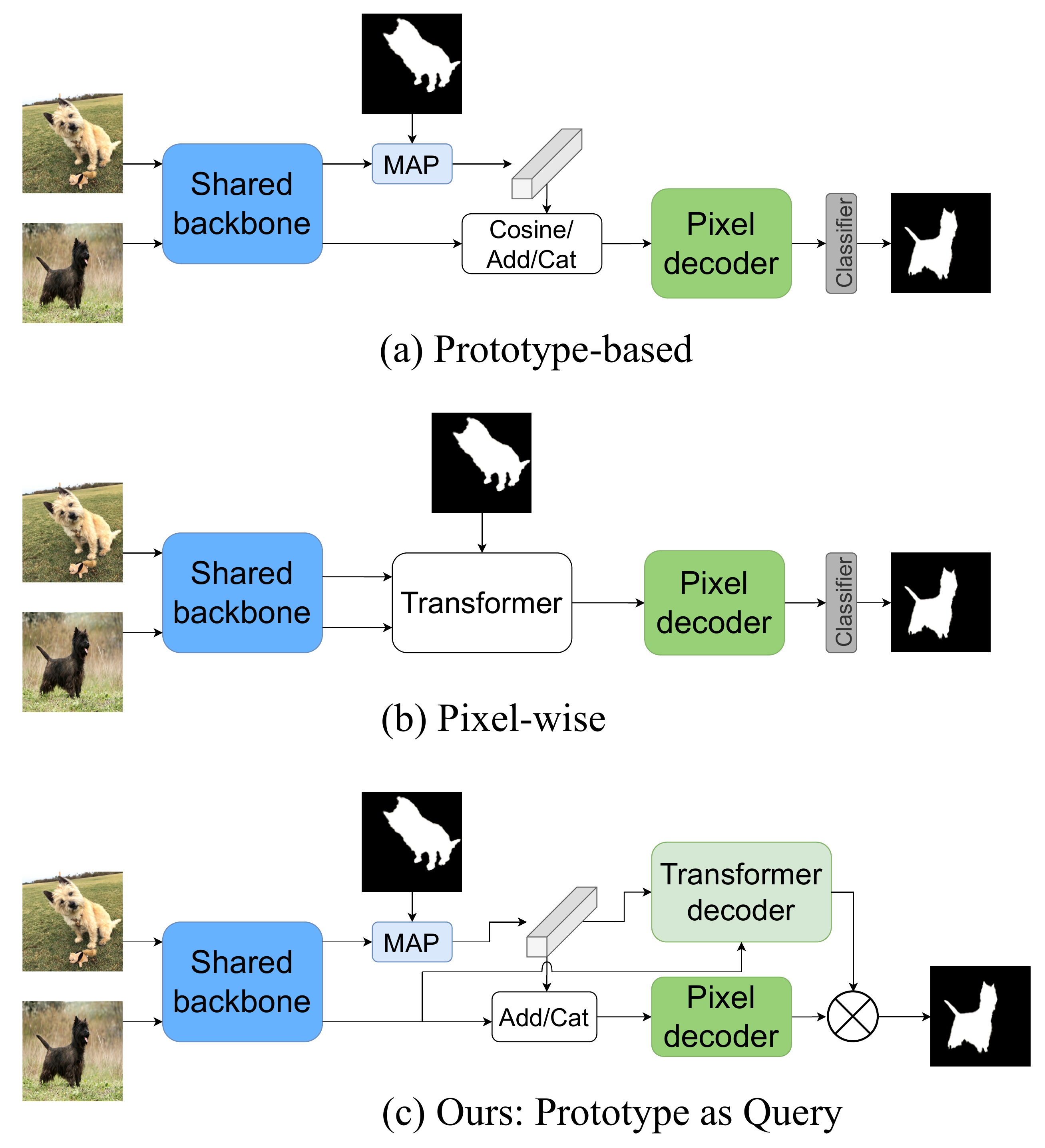}
  \caption{Comparison between existing two types of solutions and our proposed method for few-shot semantic segmentation. (a) Prototype-based method; (b) Pixel-wise method; (c) Our proposed Prototype as Query. In the figure, "MAP" represents masked average pooling operation, "Cosine" represents cosine similarity, "Add" represents element-wise sum, and "Cat" represents channel-wise concatenation.}
  \label{fig:pipeline}
\end{figure}

One of the main challenges of FSS is that the texture or appearance of the object in the query image may differ from the target object in the support examples. The key to solve FSS is to effectively exploit the dependency of the query image and the support examples. The existing methods for FSS can be mainly categorized into two groups: (1) Prototype-based methods\cite{prototype,prototypenet,cwt,panet,sg-one,pfenet}, where the semantic features of the target class in the support images extracted via a shared backbone network was abstracted into feature vectors named class-wise prototypes via class-wise average pooling or clustering. And the query features were aggregated with prototypes via element-wise sum or channel-wise concatenation, which were then enhanced by a pixel decoder module. Lastly, each pixel in the query image was classified as the target class or background. The pipeline of this type of methods is shown in Figure \ref{fig:pipeline}(a). (2) Pixel-wise methods\cite{dan,cyctr,dcama}. Instead of squeezing discriminative information into prototypes, these methods utilized all pixel-wise support information for the query predictions via computing the pixel-wise correlations between paired query and support features implemented with the attention mechanism of Transformer, as shown in Figure~\ref{fig:pipeline}(b). Although these two types of methods have demonstrated their effectiveness, they still have some limitations. First, the prototypes may lose some local and precious semantic features of the target class in the support images, which may results in segmenting the target class roughly in the query image especially for the objects with complex appearances. Second, although the pixel-wise methods have greatly boost the performance over the prototype-based methods, they suffer from heavy computation on the dot-product attention between all pixels of support features and the query features. In addition, excessively abundant pixel-wise support information may confuse the attention\cite{cyctr}.

The limitations of existing approaches motivate us to develop a simple and effective framework for solving FSS. Recently, the Query-based\footnote{To distinguish from the phrase "query" in few-shot segmentation, we use "Query" with capitalization to note the query sequence in the Transformer.} Transformer\cite{transformer2017,vit} architecture was introduced into the object detection\cite{carion2020end} and segmentation tasks\cite{xie2021segformer,strudel2021segmenter,zheng2021rethinking,cheng2021maskformer,wang2021end}, which has demonstrated a promising solution. In this architecture, the objects were represented by learnable Query embeddings, which reasoned about the relations of the object Query embeddings and the global image context. Several recent works proposed conditional Queries to make the Queries focus on the target objects, e.g., using linguistic features as conditional restriction in the referring video object segmentation task\cite{wu2022referformer}, using anchor boxes or reference points in the object detection\cite{meng2021-CondDETR}. 

Inspired by the aforementioned works, we design a Transformer-based module to better build the relations between the support examples and the query image, in parallel with the pixel decoder module. The main framework of our solution is shown in Figure \ref{fig:pipeline}(c). In the Transformer-based module, we propose to use the prototype of support examples as a conditional Query, and the extracted features of the query image are viewed as Key and Value embeddings for the input of Transformer decoder. In this manner, the semantic features of the target class in the query image can be focused by the conditional Query. The output of the Transformer-based module can be viewed as semantic-aware dynamic kernels to filter out the segmentation mask from the output of the pixel decoder module. To compensate the precious information loss caused by the support prototype, the training-free prior mask proposed in PFENet\cite{pfenet} was employed to aggregate with the support prototype and query features in our framework (not shown in Figure \ref{fig:pipeline}(c)), which revealed the pixel-wise relations between support and query features. 

To summarize, our contributions are:
\begin{itemize}
    \item We propose a simple yet effective framework for few-shot semantic segmentation, termed as ProtoFormer. The support prototype is utilized as a conditional Query to focus on the target object’s mask embedding in the query image via the Transformer decoder module. And the mask embedding is viewed as dynamic kernels to filter out the segmentation mask from the output of the pixel decoder module.
    \item We design an efficient and light-weight network, whose learnable parameters are only 0.6M, and yet yields competitive performance on benchmarks.
    \item Our solution ProtoFormer performs competitively on the benchmark of PASCAL-$5^{i}$ and achieves new state-of-the-art (SOTA) on COCO-$20^{i}$ in both 1-shot and 5-shot settings.
\end{itemize}

\section{Related Work}
\subsection{Semantic Segmentation}
Semantic segmentation, a fundamental task in computer vision, aims to classify each pixel in an image into different category labels\cite{fcn}. Performance of approaches for semantic segmentation has been significantly improved, starting from seminal work of Fully Convolutional Networks (FCNs) which replaced the fully-connected layer in a classification framework with convolutional layers\cite{long2015,fcn}. Based on the framework of FCNs, many works have been proposed through enlarging the receptive field or aggregating long-range context in the feature map. DeepLab\cite{deeplab,deeplabv3} applied dilated convolutions with different dilated rates to acquire various receptive fields of feature maps. PSPNet\cite{pspnet} utilized spatial pyramid pooling with different kernel sizes to capture contextual information of multiple scales. DANet\cite{fu2019dual} and CCNet\cite{huang2018ccnet,huang2020ccnet} used non-local blocks to exploit object contexts. Recently, SETR\cite{zheng2021rethinking}, Segmenter\cite{strudel2021segmenter} and SegFormer\cite{xie2021segformer} utilized the Transformer-based backbone to replace the traditional convolutional networks, which can better capture long-range context. Swin Transformer\cite{Liu_2021_ICCV} proposed a hierarchical architecture computed with shifted windows for general-purpose backbone, which achieved state-of-the-art performance on the semantic segmentation benchmarks. BEiT\cite{bao2022beit} introduced a self-supervised approach termed as masked image modeling to pretrain vision Transformers, which could be directly fine-tuned on the semantic segmentation task and achieved competitive results. The aforementioned approaches formulated semantic segmentation as a per-pixel classification task, MaskFormer\cite{cheng2021maskformer,cheng2021mask2former} however proposed to utilize the mask classification-based method to predict a set of binary masks, each associated with a single class label.

\subsection{Few-shot Semantic Segmentation}
Early few-shot segmentation approaches followed the metric learning framework to segment query images, where the semantic information of the target class in the support images was abstracted into prototype vectors. PL\cite{prototypenet} extended the prototype learning paradigm in few-shot learning\cite{prototype} to FSS, in which each class was represented by a prototype vector and the cosine similarity between pixels in the query features and prototypes was computed to predict the segmentation map. In PL, prototype vectors were computed by a global average pooling layer, SG-One\cite{sg-one} proposed a masked average pooling operator to compute prototype vectors for support examples. PANet\cite{panet} introduced a prototype alignment regularization between support and query images, which encouraged the prototypes generated from the queries to align well with those of the supports. Instead of computing the cosine similarity map between prototypes and query features, CANet\cite{canet} proposed a two-branch dense comparison module which performed multi-level feature comparison between supports and query, along with an iterative optimization module to refine the predicted results. A single prototype vector may cause semantic ambiguity, PMMs\cite{pmm} proposed to use an Expectation-Maximization (EM) algorithm to generate multiple prototype vectors to enforce the semantic representation. PFENet\cite{pfenet} proposed a training-free prior mask indicating the pixel-wise relation between the support and query image, along with a feature enrichment module to adaptively enrich query features with the support features. DPCN\cite{dpcn} proposed a dynamic convolution module to generate dynamic kernels from support foreground, and the convolution operations on query features were implemented with the dynamic kernels. SSP\cite{ssp} proposed a self-support matching strategy to solve the appearance discrepancy problem in FSS, where the high-confidence query predictions were abstracted to query prototypes to match query features. BAM\cite{bam} applied an additional branch (base learner) to the meta learner to segment the target objects of base classes, and the prediction results of two branches were adaptively integrated to yield precise results. NTRENet\cite{ntrenet} proposed a Non-Target Region Eliminating network to mine and eliminate background and distracting objects regions in the query image via learning a general background prototype.

Recently, some approaches exploited pixel-wise support information for FSS. PGNet\cite{pgnet} and DAN\cite{dan} utilized the graph attention network to build the relations between all foreground support pixel features and query features. HSNet\cite{hsnet} squeezed diverse features from different levels of intermediate convolutional layers and transformed to a segmentation mask with 4D convolutions. CyCTR\cite{cyctr} proposed a cycle-consistent attention mechanism to filter out possible harmful support features and made query features attend to the beneficial information of support features. DCAMA\cite{dcama} exploited all support information via multi-level pixel-wise attentions between query and support features, which treated each query pixel as a token and computed its similarities with all support feature pixels. VAT\cite{vat} introduced a cost aggregation network termed as 4D Convolutional Swin Transformer\cite{Liu_2021_ICCV} to aggregate information between query and support features, where a high-dimensional Swin Transformer was implemented by a series of small-kernel convolutions that impart local context to all pixels.

\begin{figure*}[t]
  \centering
  \includegraphics[width=1.0\linewidth]{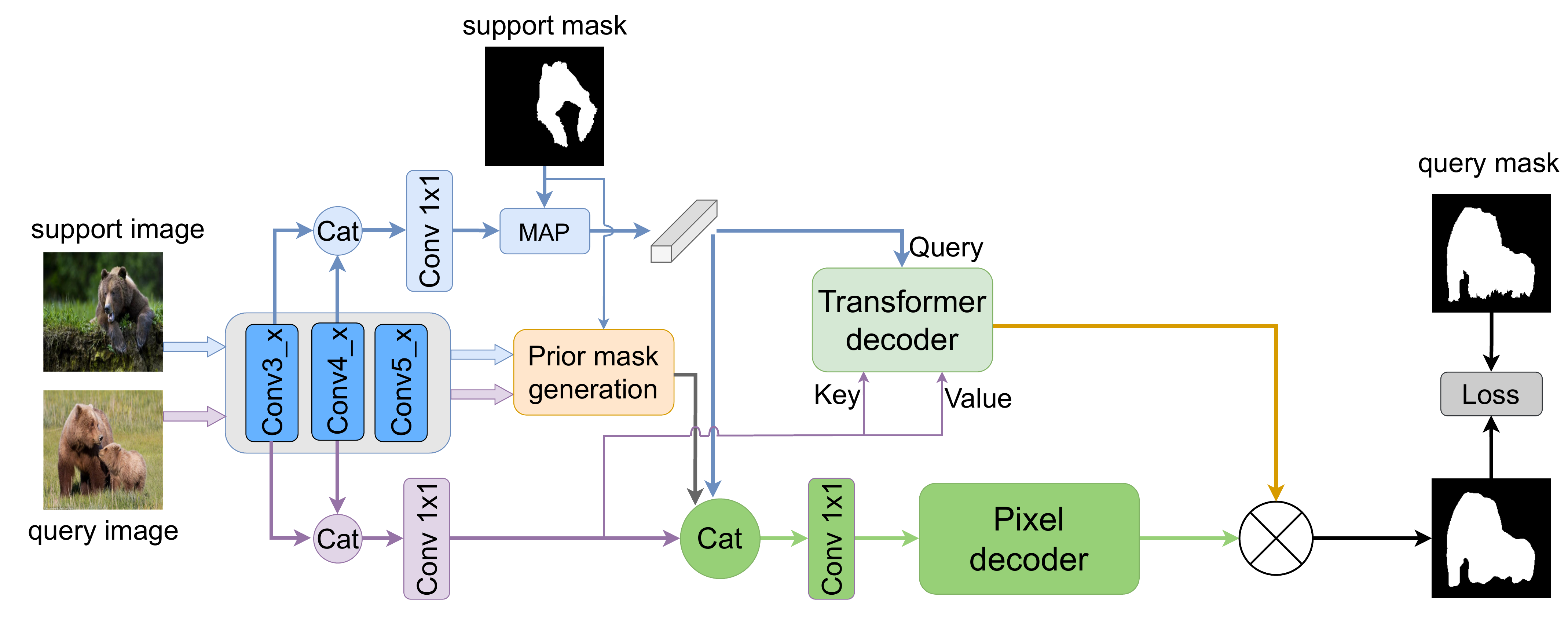}
  \caption{The overview of ProtoFormer. It is mainly composed of three parts: a shared pretrained backbone, a Transformer decoder module, and a pixel decoder module. The model takes a query image and a support image with the corresponding support mask as input and outputs the segmentation mask of the target class in the query image. In the figure, "MAP" represents masked average pooling operation, "Cat" represents channel-wise concatenation. The support features are abstracted into a prototype vector, which is view as a conditional Query for the input of Transformer decoder module. And the query features extracted from the query image are viewed as the Key and Value embeddings for the Transformer decoder module. In parallel with Transformer decoder module, the query features, the prototype vector and the prior mask are concatenated to get enriched query features via the pixel decoder module. Finally, a binary mask is acquired via a dot product between the enriched query features and the output of Transformer decoder module followed by a sigmoid activation, which is supervised by the ground truth query mask for training.}
  \label{fig:overview}
\end{figure*}

\section{Methodology}
\subsection{Problem Definition}
Few-shot semantic segmentation aims at segmenting unseen classes with only few annotated samples. Given two datasets $D_{train}$ and $D_{test}$ with class set $C_{train}$ and $C_{test}$ respectively, where $C_{train} \cap C_{support}=\emptyset$, the model trained on $D_{train}$ is expected to generalize on $D_{test}$. Current methods train models within the episode-based meta-learning paradigm, wherein both $D_{train}$ and $D_{test}$ are composed of a certain amount of randomly sampled episodes. Each episode consists of a support set $S=\{(I_s^{i}, M_s^{i})\}_{i=1}^{k}$ and a query set $Q=\{(I_q,M_q)\}$ with the same class label $c$, where $I_*$ and $M_*$ represent a raw image and its corresponding binary mask for the class label, respectively. In the meta-learning stage, the model takes the support set $S$ and the query image $I_q$ as input, and predicts the binary mask of the specific class for $I_q$, which is supervised by the ground truth binary mask $M_q$. After training on the dataset $D_{train}$, the model is directly evaluated on $D_{test}$ across all test episodes without implementing any fine-tuning stage.

\subsection{Overview}
Given a support set $S=\{(I_s^{i}\in \mathbb{R}^{3\times H \times W}, M_s^{i}\in \mathbb{R}^{1\times H \times W})\}_{i=1}^{k}$ and a query image $I_q\in \mathbb{R}^{3\times H \times W}$, we aim to predict the binary segmentation mask for $I_q$, which identifies the same class with the support examples. To this end, we propose a simple and effective framework termed as ProtoFormer, as shown in Figure~\ref{fig:overview}. We only illustrate a 1-shot setting in the framework for simplicity. ProtoFormer is mainly composed of three parts: a shared pretrained backbone, a Transformer decoder module, and a pixel decoder module. The support image $I_s$ and query image $I_q$ are input into the shared backbone network respectively, to acquire their mid-level and high-level features. The mid-level features of the support image are abstracted into a prototype vector via a masked average pooling operation associated with the support mask. The high-level features of the support and query images are used to generate the prior mask under the condition of support mask, which indicates the probability of pixels belonging to a target class. The prototype vector, prior mask and query features are concatenated and input to a pixel decoder module to further enrich the semantic features of the query image and obtain the pixel embeddings. We design a parallel module utilizing the Transformer decoder to focus on the features of target class in the query image, where the prototype vector is treated as a Query embedding and the features extracted from the query image are viewed as Key and Value embeddings. The output of Transformer decoder module can be termed as a query mask embedding. Finally, ProtoFormer predicts a binary mask via a dot product between the pixel embeddings and the mask embedding followed by a sigmoid activation. In the following subsections, we describe each part of ProtoFormer in detail.

\subsection{Pretrained backbone and pixel decoder}
\noindent \textbf{Backbone.}
In our proposed framework, we use the ImageNet\cite{deng2009} pre-trained CNN to extract features from raw images. Specifically, the dilated version of ResNet\cite{resnet} with shared weights between support and query images is utilized to extract mid-level features (conv3\_x and conv4\_x) and high-level features (conv5\_x). The mid-level and high-level features of the support image are denoted as $X_s^{c3}$, $X_s^{c4}$ and $X_s^{c5}$, respectively. And the mid-level and high-level features of the query image are denoted as $X_q^{c3}$, $X_q^{c4}$ and $X_q^{c5}$, respectively. We concatenate and yield a merged mid-level feature $X_s^m$ with $C=64$ output channels using a $1\times1$ convolution layer:
\begin{eqnarray}
    \label{eq1}
    X_s^m=\mathcal{F}_{1\times1}(Cat(X_s^{c3},X_s^{c4}))\in \mathbb{R}^{C\times \frac{H}{8} \times \frac{W}{8}},\\
    \label{eq2}
    X_q^m=\mathcal{F}_{1\times1}(Cat(X_q^{c3},X_q^{c4}))\in \mathbb{R}^{C\times \frac{H}{8} \times \frac{W}{8}}
\end{eqnarray}
where $Cat$ represents channel-wise concatenation, $\mathcal{F}_{1\times1}$ denotes the $1\times1$ convolution. 

\noindent \textbf{Prototype.}
We then abstract the mid-level feature of the support example $X_s^m$ into a prototype vector via the masked average pooling (MAP) to provide the referring class-related features:
\begin{equation}
    \label{eq3}
    p=\mathcal{F}_{pool}(X_s^m\odot \mathcal{I}(M_s))\in \mathbb{R}^{C\times 1\times 1},
\end{equation}
where $\mathcal{F}_{pool}$ is the average pooling operation, $\odot$ represents Hadamard product, and $\mathcal{I}$ is a function that reshapes $M_s$ to be the same shape as $X_s^m$ via interpolation and expansion. In the 5-shot setting, we simply take the average of five prototype vectors as a novel prototype vector\cite{pfenet}.

\noindent \textbf{Prior mask generation.}
The high-level query and support features $X_q^{c5}$ and $X_s^{c5}$ associated with the support mask $M_s$ are utilized to generate a prior mask denoted as $M_p\in \mathbb{R}^{1\times \frac{H}{8} \times \frac{W}{8}}$ indicating the pixel-wise relations between query and support features. A pixel in query feature with a high value on $M_p$ means that it is strongly related with at least on pixel of target object in the support feature. Thus, this pixel in the query image is very likely to be classified into the target class. The details of generating the prior mask can refer to \cite{pfenet}. Similarly, in the 5-shot setting, five prior masks are averaged as a novel prior mask.

\noindent \textbf{Pixel decoder.}
The prototype vector, the prior mask and query features are concatenated, and we yield a merged query feature with $c=64$ output channels using a $1\times1$ convolution layer:
\begin{equation}
    X_q=\mathcal{F}_{1\times1}(Cat(X_q^m, M_p,\mathcal{I}(p)))\in \mathbb{R}^{C\times \frac{H}{8} \times \frac{W}{8}}
\end{equation}
where $\mathcal{I}$ is a function that reshapes $p$ to be the same shape as $X_q^m$ via expansion.

To enrich the multi-scale spatial information of the query features, a pixel decoder module is designed. Any semantic segmentation decoder can be utilized here, e.g., PPM\cite{pspnet}, ASPP\cite{deeplabv3} and FEM\cite{pfenet}. In our ProtoFormer, we use FEM to be the pixel decoder module, which constructs the hierarchical relations to enrich coarse feature maps with information from the finer features via a top-down path. Details of FEM can refer to \cite{pfenet}. Thus, the new query features are obtained as:
\begin{equation}
    X_{q,new}=\mathcal{F}_{FEM}(X_q)\in \mathbb{R}^{C\times \frac{H}{8} \times \frac{W}{8}}
\end{equation}
where $\mathcal{F}_{FEM}$ represents the FEM module.

\begin{figure}[t]
  \centering
  \includegraphics[width=.7\linewidth]{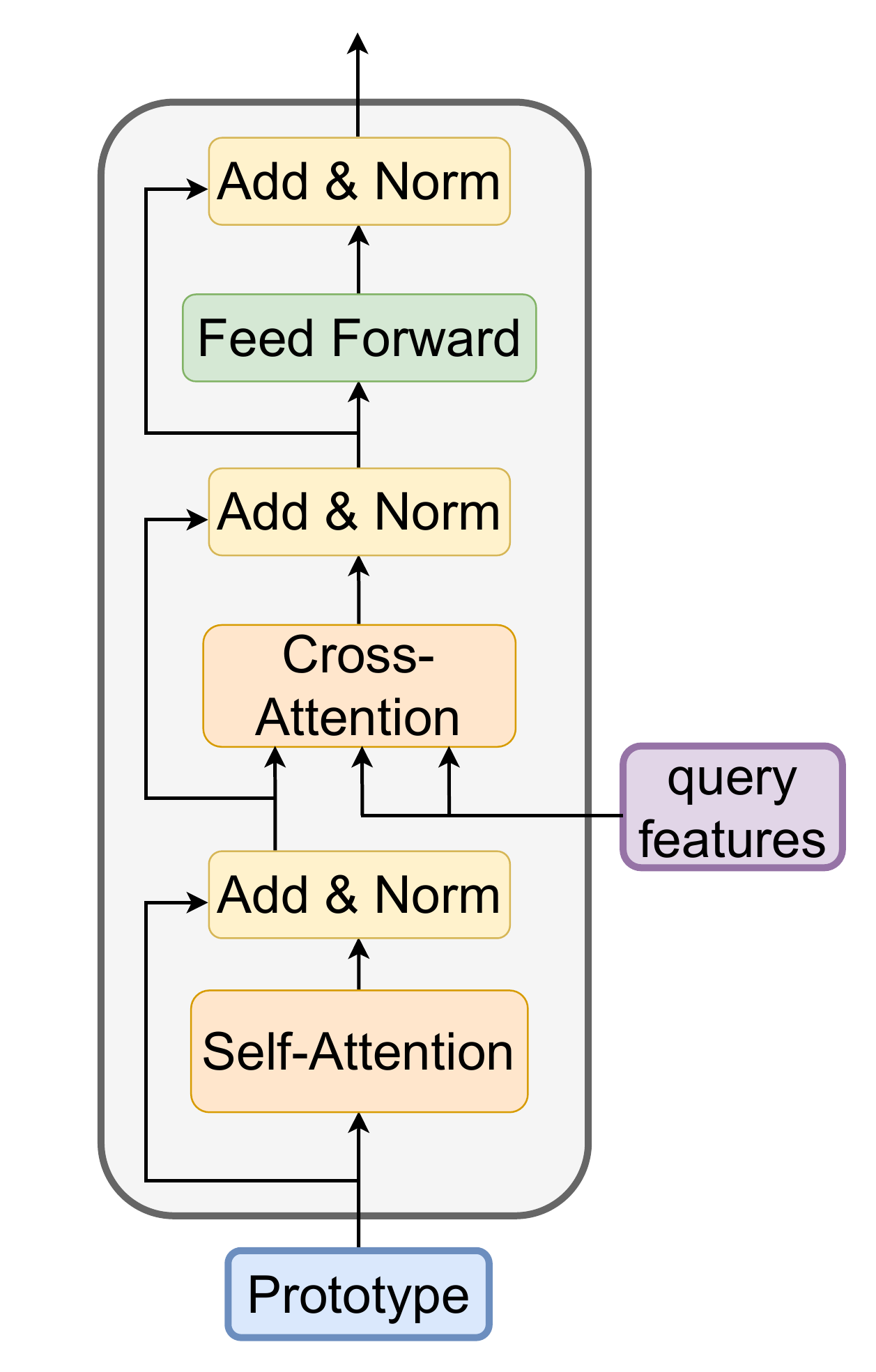}
  \caption{The architecture of Transformer decoder.}
  \label{fig:transformer}
\end{figure}

\subsection{Prototype as Query}
In parallel with the pixel decoder module, we design a Transformer decoder module to focus on the target objects in query features and produce the semantic-aware dynamic kernels. The final segmentation mask is acquired by performing dynamic convolution between the dynamic kernels and the enriched query features $X_q,new$. We use the standard Transformer decoder to implement this process. The architecture details of Transformer decoder is illustrated as Figure \ref{fig:transformer}. The prototype vector $p$ is treated as a conditional Query associated with a learnable positional encoding (not shown in the figure), which is first input to a self-attention module in a residual connection manner followed by a layer normalization and to be the Query embedding of the cross-attention module. The query features $X_q^m$ are viewed as Key and Value embeddings for the cross-attention. After implementing cross-attention, a feed-forward layer is applied. We only use one Transformer decoder layer. The output of Transformer decoder module can be viewed as a segmentation mask embedding of the target object, which we denote as $q\in \mathbb{R}^{1\times C}$.

\noindent \textbf{Segmentation head.}
To predict a binary segmentation mask, we directly make a dot-product between the segmentation mask embedding $q$ and the new query features $X_{q,new}$:
\begin{equation}
    \hat{M}=\mathcal{F}_{sigmoid}(q\otimes X_{q,new})\in \mathbb{R}^{1\times \frac{H}{8} \times \frac{W}{8}}
\end{equation}
where $\mathcal{F}_{sigmoid}$ represents a sigmoid activation.

\subsection{Training Loss}
For training our ProtoFormer, we use a dice loss between the predicted mask $\hat{M}$ and the corresponding ground truth query mask $M_q$:
\begin{equation}
    \mathcal{L}=\frac{1}{ep}\sum_{i=1}^{ep}DICE(\mathcal{I}(\hat{M}),M_q)
\end{equation}
where $ep$ is the total number of training episodes in each batch, $DICE$ represents the dice loss function, $\mathcal{I}$ represents an interpolation function that makes $\hat{M}$ to be the same spatial size as $M_q$.

\section{Experiments}
\subsection{Datasets and metrics}
\noindent \textbf{Datasets.}
We evaluate our proposed ProtoFormer on two standard FSS benchmarks: PASCAL-$5^i$\cite{shaban} and COCO-$20^i$\cite{Nguyen_2019_ICCV}. PASCAL-$5^i$ was created from PASCAL VOC 2012\cite{pascal} with extended annotations from SDS\cite{sds}, which is composed of 20 object classes. COCO-$20^i$ is a larger and more challenging benchmark created from the MSCOCO dataset\cite{coco}, which contains 80 object classes. The object categories of both PASCAL-$5^i$ and COCO-$20^i$ are evenly split into 4 folds. When testing ProtoFormer on one fold, 1,000 episodes from this fold are randomly sampled, and the remaining three folds are used to train the model for cross-validation. 

\noindent \textbf{Metrics.}
We adopt the mean intersection over union (mIoU) and foreground-background IoU (FB-IoU) as the evaluation metrics. The mIoU averages over IoU values of all classes in a fold: $mIoU=\frac{1}{N}\sum_{n=1}^{N} IoU_n$ where $N$ is the number of classes in the target fold and $IoU_n$ is the intersection over union of class $n$. FB-IoU neglects object classes and directly averages foreground and background IoU. We take the average of results on all folds as the final mIoU and FB-IoU.

\subsection{Implementation details}
All experiments are conducted on Pytorch 1.8.1. For the shared pretrained backbone, we use ResNet-50 and ResNet-101 pretrained on ImageNet and its parameters are freezed. Other layers of ProtoFormer are initialized by the default setting of Pytorch. The model is trained with the Adam optimizer with learning rate of $10^{-3}$ on PASCAL-$5^i$ for 60 epochs and COCO-$20^i$ for 30 epochs. All images are directly resized to 473$\times$ 473, and the training batch size is set to 32 and 16 for 1-shot setting and 5-shot setting, respectively. We do not use data augmentation strategies for training. The predicted binary mask is resized to the original size of images for evaluation. We average the results of 5 trails with different random seeds. Our experiments are conducted on NVIDIA RTX 3090 GPUs. Our source code and pretrained models are available at \url{https://github.com/LeileiCao/ProtoFormer}.

\begin{table*}
     \caption{Comparison with other state-of-the-art methods for 1-shot and 5-shot segmentation on PASCAL-5$^{i}$ using the mIoU and FB-IoU metrics. Numbers in bold indicate the best performance and underlined ones are the second best.}
    \label{tab:pascal}
    \begin{center}
    \scalebox{0.9}{
    \begin{tabular}{cl|cccccc|cccccc|c}
            \toprule
            \multirow{2}{*}{\shortstack{Backbone\\network}} & \multirow{2}{*}{Methods} & \multicolumn{6}{c|}{1-shot} & \multicolumn{6}{c|}{5-shot} & \# learnable \\ 
            
            & & fold0 & fold1 & fold2 & fold3 & mean & FB-IoU & fold0 & fold1 & fold2 & fold3 & mean & FB-IoU & params \\
            \midrule

            \multirow{16}{*}{ResNet50} & PANet (ICCV'19)\cite{panet}  & 44.0 & 57.5 & 50.8 & 44.0 & 49.1 & - & 55.3 & 67.2 & 61.3 & 53.2 & 59.3 & - & 23.5M \\  
            
            & PGNet (ICCV'19)\cite{pgnet}  & 56.0 & 66.9 & 50.6 & 50.4 & 56.0 & 69.9 & 57.7 & 68.7 & 52.9 & 54.6 & 58.5 & 70.5 & 17.2M \\  
            
            & PPNet (ECCV'20)\cite{ppnet}  & 48.6 & 60.6 & 55.7 & 46.5 & 52.8 & 69.2 & 58.9 & 68.3 & 66.8 & 58.0 & 63.0 & 75.8 & 31.5M \\  
            
            & PMM (ECCV'20)\cite{pmm} & 52.0 & 67.5 & 51.5 & 49.8 & 55.2 & - & 55.0 & 68.2 & 52.9 & 51.1 & 56.8 & - & - \\
        
            & PFENet (TPAMI'20)\cite{pfenet} & 61.7 & 69.5 & 55.4 & 56.3 & 60.8 & 73.3 & 63.1 & 70.7 & 55.8 & 57.9 & 61.9 & 73.9 & 10.3M \\ 
            
            & RePRI (CVPR'21)\cite{repri} & 59.8 & 68.3 & 62.1 & 48.5 & 59.7 & - & 64.6 & 71.4 & 71.1 & 59.3 & 66.6 & - & - \\ 
            
            & HSNet (ICCV'21)\cite{hsnet}  & 64.3 & 70.7 & 60.3 & 60.5 & 64.0 & 76.7 & 70.3 & 73.2 & 67.4 & \textbf{67.1} & 69.5 & 80.6 & \underline{2.5M} \\
            
            & CWT (ICCV'21)\cite{cwt} & 56.3 & 62.0 & 59.9 & 47.2 & 56.4 & - & 61.3 & 68.5 & 68.5 & 56.6 & 63.7 & - & - \\
            
            & CyCTR (NeurIPS'21)\cite{cyctr} & 65.7 & 71.0 & 59.5 & 59.7 & 64.0 & - & 69.3 & 73.5 & 63.8 & 63.5 & 67.5 & - & 15.4M \\
            
            & BAM (CVPR'22)\cite{bam} & \underline{69.0} & \underline{73.6} & \underline{67.6} & \underline{61.1} & \textbf{67.8} & \textbf{79.7} & 70.6 & \underline{75.1} & \underline{70.8} & \underline{67.0} & \textbf{70.9} & \textbf{82.2} & 26.7M \\
            
            & NTRENet (CVPR'22)\cite{ntrenet} & 65.4 & 72.3 & 59.4 & 59.8 & 64.2 & 77.0 & 66.2 & 72.8 & 61.7 & 62.2 & 65.7 & 78.4 &19.9M \\
            
            & DPCN (CVPR'22)\cite{dpcn} & 65.7 & 71.6 & \textbf{69.1} & 60.6 & 66.7 & \underline{78.0} & 70.0 & 73.2 & \textbf{70.9} & 65.5 & 69.9 & 80.7 & - \\
            
            & VAT (ECCV'22)\cite{vat} & 67.6 & 72.0 & 62.3 & 60.1 & 65.5 & 77.8 & \underline{72.4} & 73.6 & 68.6 & 65.7 & \underline{70.1} & 80.9 & 3.2M \\
            & SSP (ECCV'22) \cite{ssp} & 60.5 & 67.8 & 66.4 & 51.0 & 61.4 & - & 68.0 & 72.0 & 74.8 & 60.2 & 68.8 & - & 8.7M \\
            
            & DCAMA (ECCV'22)\cite{dcama} & 67.5 & 72.3 & 59.6 & 59.0 & 64.6 & 75.7 & 70.5 & 73.9 & 63.7 & 65.8 & 68.5 & 79.5 & 47.7M \\
            
            & IPMT (NeurIPS'22)\cite{ipmt} & \textbf{72.8} & \textbf{73.7} & 59.2 & \textbf{61.6} & \underline{66.8} & 77.1 & \textbf{73.1} & 74.7 & 61.6 & 63.4 & 68.2 & \underline{81.4} & - \\
            \cline{2-15} \\[-2.0ex]
            & ProtoFormer (Ours) & 65.9 & 72.5 & 55.9 & 58.1 & 63.1 &72.6 & 71.4 & \textbf{75.2} & 57.5 & 65.7 & 67.4 & 77.1 & \textbf{0.6M} \\
            \midrule
            
            \multirow{13}{*}{ResNet101} & FWB (ICCV'19)\cite{fwb}  & 51.3 & 64.5 & 56.7 & 52.2 & 56.2 & - & 54.8 & 67.4 & 62.2 & 55.3 & 59.9 & - & 43.0M \\
            
            & PPNet (ECCV'20)\cite{ppnet}   & 52.7 & 62.8 & 57.4 & 47.7 & 55.2 & 70.9 & 60.3 & 70.0 & 69.4 & 60.7 & 65.1 & 77.5 & 50.5M \\  
            
            & DAN (ECCV'20)\cite{dan}    & 54.7 & 68.6 & 57.8 & 51.6 & 58.2 & 71.9 & 57.9 & 69.0 & 60.1 & 54.9 & 60.5 & 72.3 & - \\ 
            
            & PFENet (TPAMI'20)\cite{pfenet} & 60.5 & 69.4 & 54.4 & 55.9 & 60.1 & 72.9 & 62.8 & 70.4 & 54.9 & 57.6 & 61.4 & 73.5 & 10.3M \\
            
            & RePRI (CVPR'21)\cite{repri} & 59.6 & 68.6 & 62.2 & 47.2 & 59.4 & - & 66.2 & 71.4 & 67.0 & 57.7 & 65.6 & - & - \\ 
            
            & HSNet (ICCV'21)\cite{hsnet}  & 67.3 & 72.3 & 62.0 & \underline{63.1} & \underline{66.2} & 77.6 & 71.8 & 74.4 & 67.0 & \underline{68.3} & 70.4 & 80.6 & \underline{2.5M} \\
            
            & CWT (ICCV'21)\cite{cwt} & 56.9 & 65.2 & 61.2 & 48.8 & 58.0 & - & 62.6 & 70.2 & \underline{68.8} & 57.2 & 64.7 & - & - \\
            
            & CyCTR (NeurIPS'21)\cite{cyctr} & 69.3 & \underline{72.7} & 56.5 & 58.6 & 64.3 & 73.0 & 73.5 & 74.0 & 58.6 & 60.2 & 66.6 & 75.4 & 15.4M \\
            
            & NTRENet (CVPR'22) & 65.5 & 71.8 & 59.1 & 58.3 & 63.7 & 75.3 & 67.9 & 73.2 & 60.1 & 66.8 & 67.0 & 78.2 & 19.9M \\
            
            & VAT (ECCV'22)\cite{vat} & \underline{70.0} & 72.5 & \underline{64.8} & \textbf{64.2} & \textbf{67.9} & \textbf{79.6} & \underline{75.0} & 75.2 & 68.4 & \textbf{69.5} & \underline{72.0} & \textbf{83.2} & 3.3M \\
            
            & SSP (ECCV'22)\cite{ssp} & 63.7 & 70.1 & \textbf{66.7} & 55.4 & 64.0 & - & 70.3 & \underline{76.3} & \textbf{77.8} & 65.5 & \textbf{72.5} & - & 27.7M \\
            
            & DCAMA (ECCV'22)\cite{dcama} & 65.4 & 71.4 & 63.2 & 58.3 & 64.6 & 77.6 & 70.7 & 73.7 & 66.8 & 61.9 & 68.3 & \underline{80.8} &47.7M \\
            
            & IPMT (NeurIPS'22)\cite{ipmt} & \textbf{71.6} & \textbf{73.5} & 58.0 & 61.2 & 66.1 & \underline{78.5} & \textbf{75.3} & \textbf{76.9} & 59.6 & 65.1 & 69.2 & 80.3 & - \\
            
            \cline{2-15} \\[-2.0ex]
            & ProtoFormer (Ours) & 67.0 & 72.2 & 55.0 & 58.4 & 63.2 & 72.6 & 71.3 & 75.8 & 55.3 & 66.1 & 67.0 & 76.3 & \textbf{0.6M} \\
            
            \bottomrule
    \end{tabular}
    }
    \vspace{-2.0mm}
    \end{center}
\end{table*}

\begin{table*}
     \caption{Comparison with other state-of-the-art methods for 1-shot and 5-shot segmentation on COCO-20$^{i}$ using the mIoU and FB-IoU metrics. Numbers in bold indicate the best performance and underlined ones are the second best.}
    \label{tab:coco}
    \begin{center}
    \scalebox{0.9}{
    \begin{tabular}{cl|cccccc|cccccc|c}
            \toprule
            \multirow{2}{*}{\shortstack{Backbone\\network}} & \multirow{2}{*}{Methods} & \multicolumn{6}{c|}{1-shot} & \multicolumn{6}{c|}{5-shot} & \# learnable \\ 
            
            & & fold0 & fold1 & fold2 & fold3 & mean & FB-IoU & fold0 & fold1 & fold2 & fold3 & mean & FB-IoU & params \\
            \midrule

            \multirow{14}{*}{ResNet50} & PPNet (ECCV'20)\cite{ppnet} & 28.1 & 30.8 & 29.5 & 27.7 & 29.0 & - & 39.0 & 40.8 & 37.1 & 37.3 & 38.5 & - & 31.5M \\  
            
            & PMM (ECCV'20)\cite{pmm} & 29.3 & 34.8 & 27.1 & 27.3 & 29.6 & - & 33.0 & 40.6 & 30.3 & 33.3 & 34.3 & - & - \\
            
            & PFENet (TPAMI'20)\cite{pfenet} & 36.5 & 38.6 & 34.5 & 33.8 & 35.8 & - & 36.5 & 43.3 & 37.8 & 38.4 & 39.0 & - & 10.3M \\ 
            
            & RePRI (CVPR'21)\cite{repri} & 32.0 & 38.7 & 32.7 & 33.1 & 34.1 & - & 39.3 & 45.4 & 39.7 & 41.8 & 41.6 & - & - \\ 
            
            & HSNet (ICCV'21)\cite{hsnet}    & 36.3 & 43.1 & 38.7 & 38.7 & 39.2 & 68.2 & 43.3 & 51.3 & 48.2 & 45.0 & 46.9 & 70.7 & \underline{2.5M} \\
            
             & CWT (ICCV'21)\cite{cwt} & 32.2 & 36.0 & 31.6 & 31.6 & 32.9 & - & 40.1 & 43.8 & 39.0 & 42.4 & 41.3 & - & - \\
            
            & CyCTR (NeurIPS'21)\cite{cyctr} & 38.9 & 43.0 & 39.6 & 39.8 & 40.3 & - & 41.1 & 48.9 & 45.2 & 47.0 & 45.6 & - & 15.4M \\
            
            & BAM (CVPR'22)\cite{bam} & \textbf{43.4} & \textbf{50.6} & \textbf{47.5} & \underline{43.4} & \textbf{46.2} & - & \textbf{49.3} & \underline{54.2} & \underline{51.6} & \underline{49.6} & \underline{51.2} & - & 26.7M \\
            
            & NTRENet (CVPR'22)\cite{ntrenet} & 36.8 & 42.6 & 39.9 & 37.9 & 39.3 & 68.5 & 38.2 & 44.1 & 40.4 & 38.4 & 40.3 & 69.2 & 19.9M \\
            
            & DPCN (CVPR'22)\cite{dpcn} & 42.0 & 47.0 & 43.3 & 39.7 & 43.0 & 63.2 & 46.0 & 54.9 & 50.8 & 47.4 & 49.8 & 67.4 & - \\
            
            & VAT (ECCV'22)\cite{vat} & 39.0 & 43.8 & 42.6 & 39.7 & 41.3 & 68.8 & 44.1 & 51.1 & 50.2 & 46.1 & 47.9 & \underline{72.4} & 3.2M \\
            
            & SSP (ECCV'22)\cite{ssp}& 35.5 & 39.6 & 37.9 & 36.7 & 37.4 & - & 40.6 & 47.0 & 45.1 & 43.9 & 44.1 & - & 8.7M \\
            
            & DCAMA (ECCV'22)\cite{dcama} & 41.9 & 45.1 & 44.4 & 41.7 & 43.3 & \underline{69.5} & 45.9 & 50.5 & 50.7 & 46.0 & 48.3 & 71.7 &47.7M \\
            
            & IPMT (NeurIPS'22)\cite{ipmt} & 41.4 & 45.1 & 45.6 & 40.0 & 43.0 & - & 43.5 & 49.7 & 48.7 & 47.9 & 47.5 & - & - \\
            \cline{2-15} \\[-2.0ex]
            & ProtoFormer (Ours) & \underline{42.4} & \underline{48.5} & \underline{46.3} & \textbf{45.5} & \underline{45.7} & \textbf{69.6} & \underline{48.1} & \textbf{57.8} & \textbf{55.0} & \textbf{52.7} & \textbf{53.4} & \textbf{73.3} & \textbf{0.6M} \\
            \midrule
            
            \multirow{10}{*}{ResNet101} & FWB (ICCV'19)\cite{fwb} & 17.0 & 18.0 & 21.0 & 28.9 & 21.2 & - & 19.1 & 21.5 & 23.9 & 30.1 & 23.7 & - & 43.0M \\
            
            & PFENet (TPAMI'20)\cite{pfenet} & 36.8 & 41.8 & 38.7 & 36.7 & 38.5 & 63.0 & 40.4 & 46.8 & 43.2 & 40.5 & 42.7 & 65.8 & 10.3M \\
            
            & SCL (CVPR'21)\cite{scl} & 36.4 & 38.6 & 37.5 & 35.4 & 37.0 & - & 38.9 & 40.5 & 41.5 & 38.7 & 39.9 & - & - \\
            
            & SAGNN (CVPR'21)\cite{sagnn} & 36.1 & 41.0 & 38.2 & 33.5 & 37.2 & - & 40.9 & 48.3 & 42.6 & 38.9 & 42.7 & - & - \\
            
            & HSNet (ICCV'21)\cite{hsnet} & 37.2 & 44.1 & 42.4 & \underline{41.3} & 41.2 & 69.1 & 45.9 & 53.0 & 51.8 & 47.1 & 49.5 & 72.4 & \underline{2.5M} \\
            
            & CWT (ICCV'21)\cite{cwt} & 30.3 & 36.6 & 30.5 & 32.2 & 32.4 & - & 38.5 & 46.7 & 39.4 & 43.2 & 42.0 & - & - \\
            
            & NTRENet (CVPR'22)\cite{ntrenet} & 38.3 & 40.4 & 39.5 & 38.1 & 39.1 & 67.5 & 42.3 & 44.4 & 44.2 & 41.7 & 43.2 & 69.6 & 19.9M \\
            
            & SSP (ECCV'22)\cite{ssp} & 39.1 & 45.1 & 42.7 & 41.2 & 42.0 & - & 47.4 & 54.5 & 50.4 & \underline{49.6} & 50.2 & - & 27.7M \\
            
            & DCAMA (ECCV'22)\cite{dcama} & \underline{41.5} & \underline{46.2} & \underline{45.2} & \underline{41.3} & \underline{43.5} & \underline{69.9} & \underline{48.0} & \underline{58.0} & \underline{54.3} & 47.1 & \underline{51.9} & \underline{73.3} & 47.7M \\
            
            & IPMT (NeurIPS'22)\cite{ipmt} & 40.5 & 45.7 & 44.8 & 39.3 & 42.6 & - & 45.1 & 50.3 & 49.3 & 46.8 & 47.9 & - & - \\
            \cline{2-15} \\[-2.0ex]
            & ProtoFormer (Ours) & \textbf{42.9} & \textbf{50.7} & \textbf{48.4} & \textbf{45.8} & \textbf{47.0} & \textbf{70.0} & \textbf{49.6} & \textbf{59.7} & \textbf{56.4} & \textbf{53.0} & \textbf{54.7} & \textbf{74.6} & \textbf{0.6M} \\
            \bottomrule
    \end{tabular}
    }
    \vspace{-2.0mm}
    \end{center}
\end{table*}

\begin{table*}
     \caption{Ablation study on the effect of our proposed prototype as Query for 1-shot and 5-shot segmentation on PASCAL-$5^i$ and COCO-20$^{i}$ using the mIoU and FB-IoU metrics. Numbers in bold indicate the best performance.}
    \label{tab:ab_baseline}
    \begin{center}
    \scalebox{0.7}{
    \begin{tabular}{cl|llllll|llllll|c}
            \toprule
            \multirow{2}{*}{Datasets} & \multirow{2}{*}{Methods} & \multicolumn{6}{c|}{1-shot} & \multicolumn{6}{c|}{5-shot} & \# learnable \\ 
            
            & & fold0 & fold1 & fold2 & fold3 & mean & FB-IoU & fold0 & fold1 & fold2 & fold3 & mean & FB-IoU & params \\
            \midrule

            \multirow{2}{*}{PASCAL-$5^i$} & Baseline & 65.1 & 71.6 & 55.4 & 56.8 & 62.2 & 71.9 & 71.0 & 74.4 & 55.6 & 64.6 & 66.4 & 76.4 & \textbf{0.58M} \\  
            
            & ProtoFormer (Ours) & \textbf{65.9}\textcolor{red}{(+0.8)} & \textbf{72.5}\textcolor{red}{(+0.9)} & \textbf{55.9}\textcolor{red}{(+0.5)} & \textbf{58.1}\textcolor{red}{(+1.3)} & \textbf{63.1}\textcolor{red}{(+0.9)} &\textbf{72.6}\textcolor{red}{(+0.7)} & \textbf{71.4}\textcolor{red}{(+0.4)} & \textbf{75.2}\textcolor{red}{(+0.8)} & \textbf{57.5}\textcolor{red}{(+1.9)} & \textbf{65.7}\textcolor{red}{(+1.1)} & \textbf{67.4}\textcolor{red}{(+1.0)} & \textbf{77.1}\textcolor{red}{(+0.7)} & 0.61M \\
            \midrule
            
            \multirow{2}{*}{COCO-$20^i$} & Baseline & 42.2 & 46.9 & 45.4 & 44.7 & 44.8 & 69.0 & 47.8 & 56.6 & 54.5 & 51.5 & 52.6 & 72.9 & \textbf{0.58M} \\
            
            & ProtoFormer (Ours) &\textbf{42.4}\textcolor{red}{(+0.2)} & \textbf{48.5}\textcolor{red}{(+1.6)} & \textbf{46.3}\textcolor{red}{(+0.9)} & \textbf{45.5}\textcolor{red}{(+0.8)} & \textbf{45.7}\textcolor{red}{(+0.9)} & \textbf{69.6}\textcolor{red}{(+0.6)} & \textbf{48.1}\textcolor{red}{(+0.3)} & \textbf{57.8}\textcolor{red}{(+1.2)} & \textbf{55.0}\textcolor{red}{(+0.5)} & \textbf{52.7}\textcolor{red}{(+1.2)} & \textbf{53.4}\textcolor{red}{(+0.8)} & \textbf{73.3}\textcolor{red}{(+0.4)} & 0.61M \\
            \bottomrule
    \end{tabular}
    }
    \vspace{-2.0mm}
    \end{center}
\end{table*}

\begin{figure*}[t]
  \centering
  \includegraphics[width=1.0\linewidth]{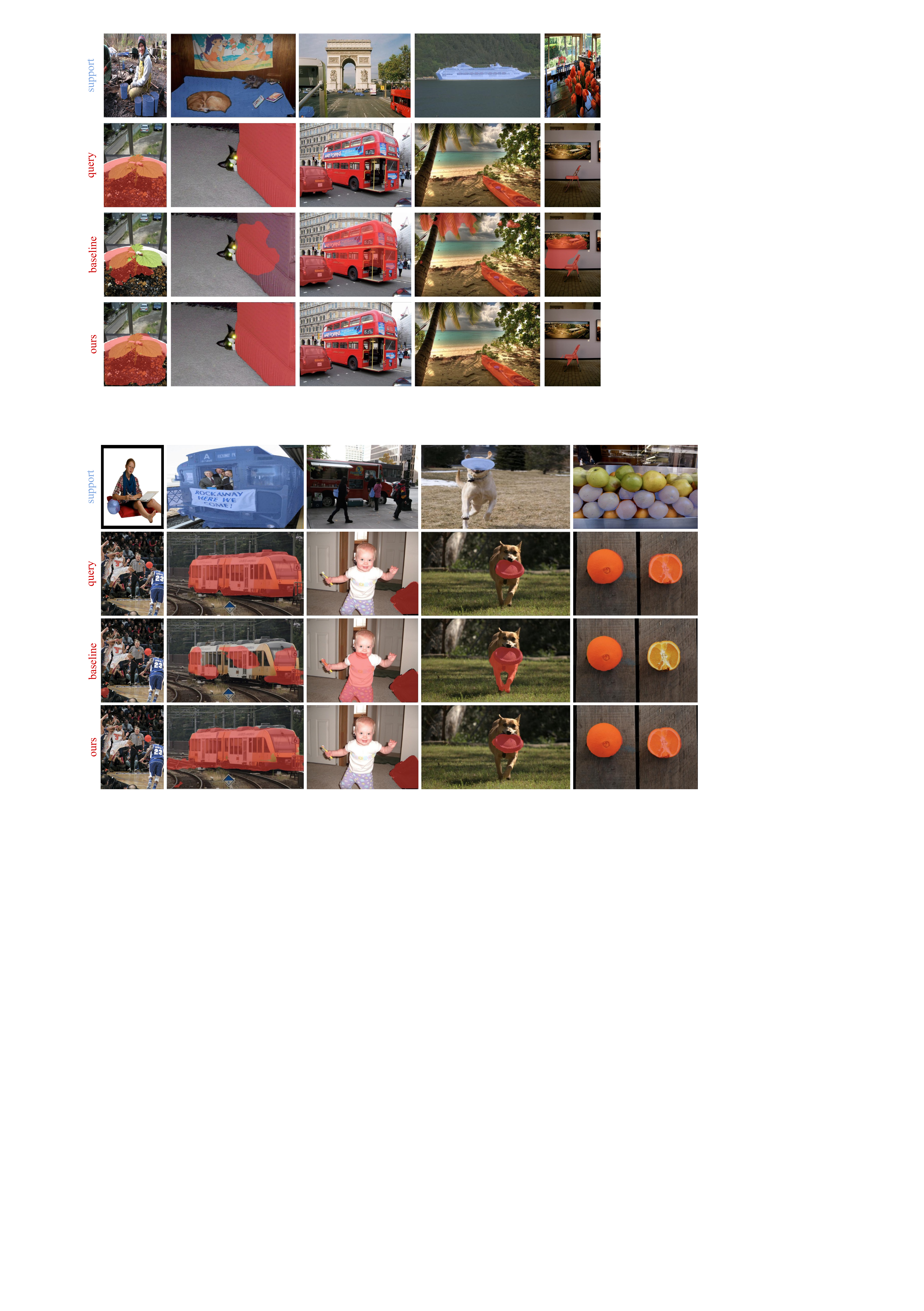}
  \caption{Qualitative results of our proposed model ProtoFormer and the baseline method for 1-shot segmentation on PASCAL-$5^i$.}
  \label{fig:pascal}
\end{figure*}

\begin{figure*}[t]
  \centering
  \includegraphics[width=1.0\linewidth]{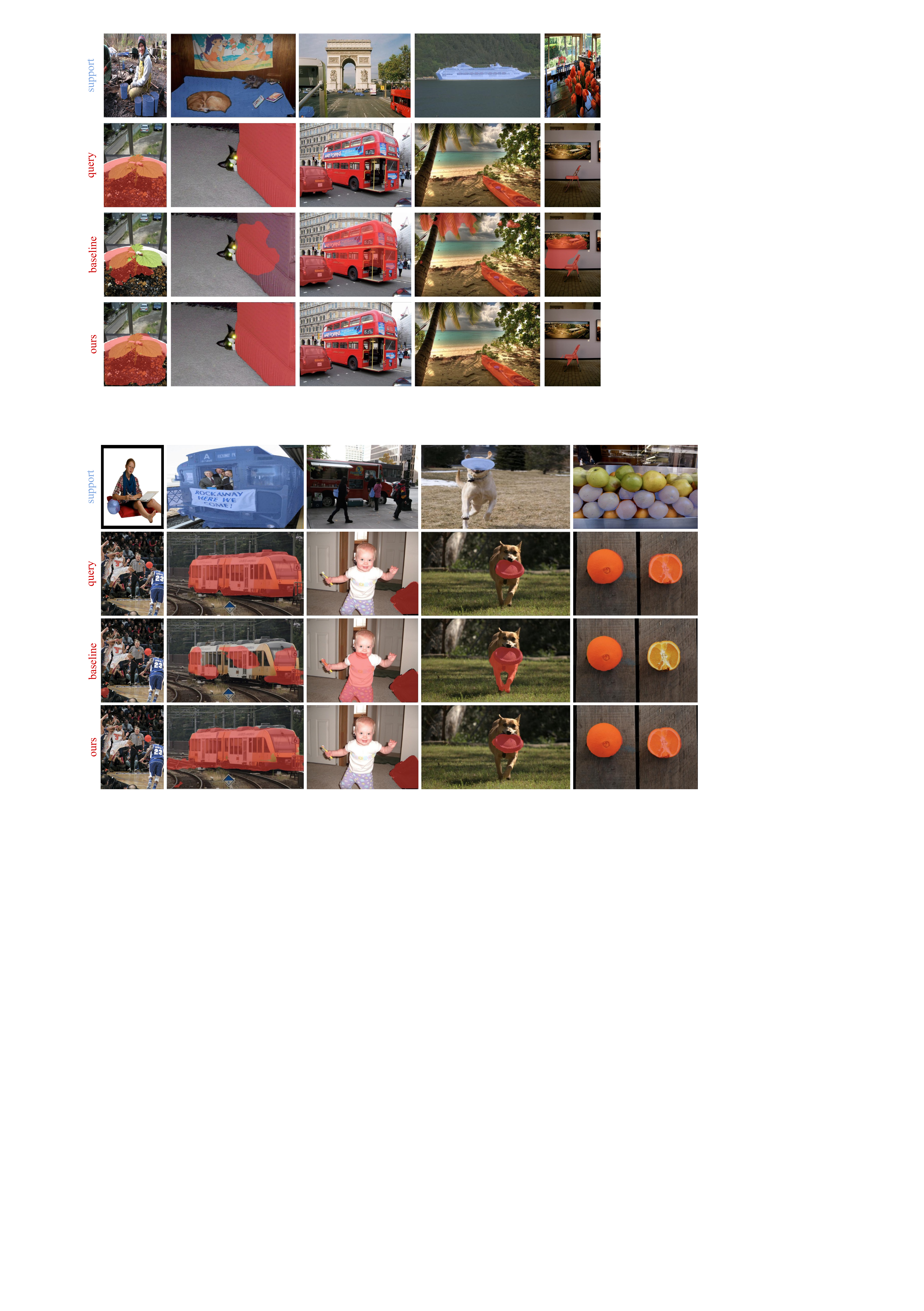}
  \caption{Qualitative results of our proposed model ProtoFormer and the baseline method for 1-shot segmentation on COCO-$20^i$.}
  \label{fig:coco}
\end{figure*}

\begin{table}
    \caption{Ablation study on the effect of varying the number of output feature channels for 1-shot on PASCAL-$5^i$ using the mIoU and FB-IoU metrics. Numbers in bold indicate the best performance.}
    \label{tab:ab_dim}
    \begin{center}
    \scalebox{0.8}{
    \begin{tabular}{l|cccccc|c}
            \toprule
              \multirow{2}{*}{dim} & \multicolumn{6}{c|}{1-shot} & \# learnable \\ 
            
             & fold0 & fold1 & fold2 & fold3 & mean & FB-IoU &  params \\
            \midrule

           16 & 63.9 & 69.4 & 53.3 & 54.7 & 60.3 & 70.4 & \textbf{0.07M} \\  
            
            32 & \textbf{66.3} & 72.1 & 55.4 & 57.3 & 62.8 & 72.4 & 0.20M \\ 
            
            64 (Ours) & 65.9 & \textbf{72.5} & \textbf{55.9} & 58.1 & \textbf{63.1} & \textbf{72.6} & 0.61M \\
            
            128 & 65.4 & 72.3 & \textbf{55.9} & 58.0 & 62.9 & \textbf{72.6} & 2.07M \\
            
            256 & 64.6 & 72.1 & 55.5 & \textbf{58.4} & 62.7 & 72.5 & 7.49M \\
            \bottomrule
    \end{tabular}
    }
    \vspace{-2.0mm}
    \end{center}
\end{table}

\begin{table}
     \caption{Ablation study on the effect of varying the number of Transformer decoder layers for 1-shot segmentation on PASCAL-$5^i$ using the mIoU and FB-IoU metrics. Numbers in bold indicate the best performance.}
    \label{tab:ab_decoder}
    \begin{center}
    \scalebox{0.8}{
    \begin{tabular}{l|cccccc|c}
            \toprule
              \multirow{2}{*}{layers} & \multicolumn{6}{c|}{1-shot} & \# learnable \\ 
            
             & fold0 & fold1 & fold2 & fold3 & mean & FB-IoU &  params \\
            \midrule
            
            1 (Ours) & 65.9 & \textbf{72.5} & \textbf{55.9} & \textbf{58.1} & \textbf{63.1} & 72.6 & \textbf{0.61M} \\
            
            2 & \textbf{66.1} & 72.3 & 55.8 & 57.1 & 62.8 & \textbf{73.0} & 0.65M \\
            
            4 & 65.3 & 72.4 & 55.8 & 58.1 & 62.9 & 72.5 & 0.72M \\
            \bottomrule
    \end{tabular}
    }
    \vspace{-2.0mm}
    \end{center}
\end{table}

\subsection{Comparison with State-of-the-Arts}
In Table \ref{tab:pascal} and Table \ref{tab:coco}, we compare our model ProtoFormer with some state-of-the-art methods on PACSAL-$5^{i}$ and COCO-20$^{i}$ respectively. We report the mIoU and FB-IoU under both 1-shot and 5-shot settings, and we only show the final FB-IoU value since of limited space. Results of other methods are taken from the original papers.

\noindent \textbf{PASCAL-$5^{i}$.}
As shown in Table \ref{tab:pascal}, our method ProtoFormer under the ResNet50 backbone outperforms PFENet by 2.3\% and 5.5\% of the mean mIoU in the 1-shot and 5-shot settings, respectively. Although our approach does not achieve the new state-of-the-art performance on this dataset,  it is still very competitive especially on fold0 and fold1 for 1-shot and 5-shot settings. The proposed Transformer decoder module in our method does not improve performance too much on fold2 comparing with PFENet. Replacing the backbone using ResNet101 does not bring improvement of ProtoFormer's performance, which also occurs on some previous SOTA methods, e.g., DCAMDA, PFENet. It is worth noting that our ProtoFormer performs competitively with the fewest learnable parameters (only 0.6M). 

\noindent \textbf{COCO-20$^{i}$.}
This benchmark contains more variant objects in a query image, which greatly challenge the generalization ability of the trained models for FSS. As shown in Table \ref{tab:coco}, our method ProtoFormer achieves new state-of-the-art performance on COCO-$20^i$ with outperforming a large margin comparing with the previous SOTA method BAM (under the ResNet50 backbone for 5-shot segmentation) and DCAMA (under the ResNet101 babckbone). Specifically, although BAM outperforms our ProtoFormer by 0.5\% of mIoU for 1-shot segmentation, yet suffers considerable disadvantages of 2.2\% of the mean mIoU for 5-shot segmentation. With the ResNet101 backbone, our ProtoFormer surpasses the previous SOTA method DCAMA by 3.5\% and 2.8\% for 1-shot and 5-shot segmentation, respectively. In addition, our ProtoFormer achieves new state-of-the-art performance using the FB-IoU metric for both 1-shot and 5-shot segmentation. It is worth noting that our ProtoFormer achieves the SOTA performance with the fewest learnable parameters (only 0.6M), which is much less than the previous SOTA method DCAMA.

\subsection{Ablation Study}
\noindent \textbf{Prototype as Query.}
In Table \ref{tab:ab_baseline}, we verify the effect of our proposed prototype as Query for 1-shot and 5-shot segmentation on PASCAL-$5^i$ and COCO-20$^{i}$ using the mIoU and FB-IoU metrics. We construct a baseline model which drops the Transformer decoder module and the pixel decoder directly follows a 1$\times$1 convolutional layer to predict a binary segmentation mask. All models use ResNet50 as backbone. We observe that our ProtoFormer improves the baseline 0.9\% and 1.0\% of the mean mIoU for 1-shot and 5-shot segmentation on PASCAL-$5^i$ respectively, at the cost of increasing 0.03M learnable parameters. The improvement on COCO-20$^{i}$ also verifies the effectiveness of our proposed idea prototype as Query. Figure \ref{fig:pascal} and Figure  \ref{fig:coco} show the qualitative results of ProtoFormer and the baseline method with the ResNet50 backbone for 1-shot segmentation on PASCAL-$5^i$ and COCO-20$^{i}$ respectively. As shown in the first two columns in Figure \ref{fig:pascal}, the baseline method only predicts a part of the target object, yet misclassifies some areas belonging to background as the target object in the last three columns. The missing classification of baseline method also occurs in Figure \ref{fig:coco}.

\noindent \textbf{Number of output feature channels.}
Table \ref{tab:ab_dim} shows the ablation study on the effect of varying the number of output feature channels for 1-shot segmentation on PASCAL-$5^i$ using the mIoU and FB-IoU metrics. ResNet50 is used as the backbone. We observe that increasing the number of channels does not improve performance of the model. It is interesting to see that even with 16 channels ProtoFormer can still achieve 60.3\% of the mean mIoU.

\noindent \textbf{Number of Transformer decoder layers.}
Table \ref{tab:ab_decoder} shows the ablation study on the effect of varying the number of Transformer decoder layers for 1-shot segmentation on PASCAL-$5^i$ using the mIoU and FB-IoU metrics. ResNet50 is used as the backbone. The results illustrate that ProtoFormer is not sensitive to the number of Transformer decoder layers. Even with 1 Transformer decoder layer ProtoFormer has achieved the best result of the mean mIoU in the table.

\section{Conclusion}
In this paper, we proposed a simple and effective framework for few-shot semantic segmentation termed as ProtoFormer, from a new perspective. In our method, the prototype of support examples is treated as a conditional Query and the query features are viewed as Key and Value embeddings for the input of a Transformer decoder layer. In this manner, the semantic features of the target class in the query image can be focused and the output of Transformer decoder can be viewed as semantic-aware dynamic kernels to filter out the segmentation mask from the enriched query features. Experimental results verify that our proposed solution ProtoFormer performs competitvely on PASCAL-5$^i$ and achieves a new state-of-the-art on COCO-20$^i$. We hope this proposed prototype as Query can motivate researchers to deeply exploit the effect of prototype of support examples.


\ifCLASSOPTIONcaptionsoff
  \newpage
\fi

\bibliographystyle{IEEEtran}
\bibliography{bare_jrnl}


\end{document}